\documentclass[mlabstract]{jmlr}

\usepackage[hpos=300px,vpos=70px]{draftwatermark}
\SetWatermarkText{\test}
\SetWatermarkScale{1}
\SetWatermarkAngle{0}

\usepackage{longtable}%

\usepackage{booktabs}
\usepackage[load-configurations=version-1]{siunitx} %

\usepackage{amsmath}
\usepackage{graphicx}
\usepackage{tikz}
\usepackage{tikz-cd}
\usepackage{listings}
\usepackage{wrapfig}
\usepackage{enumitem}

\newcommand{\CodeHLg}[1]{%
  \begingroup
  \setlength{\fboxsep}{0pt}%
  \rlap{\colorbox{green!20}{\makebox[\linewidth][l]{#1\strut}}}#1%
  \endgroup
}

\newcommand{\CodeHLr}[1]{%
  \begingroup
  \setlength{\fboxsep}{0pt}%
  \rlap{\colorbox{red!20}{\makebox[\linewidth][l]{#1\strut}}}#1%
  \endgroup
}

\lstdefinestyle{mypython}{
  language=Python,
  basicstyle=\ttfamily\small,
  keepspaces=true,
  columns=fullflexible,
  showstringspaces=false,
  moredelim=**[is][\CodeHLg]{@g}{@},
  moredelim=**[is][\CodeHLr]{@r}{@},
}

\theorembodyfont{\upshape}
\theoremheaderfont{\scshape}
\theorempostheader{:}
\theoremsep{\newline}

\jmlrvolume{}
\firstpageno{1}

\jmlryear{2025}
\jmlrworkshop{Symmetry and Geometry in Neural Representations}

\title[Equivariance by Local Canonicalization: A Matter of Representation]{\vspace*{-0.1cm}Equivariance by Local Canonicalization: \\ A Matter of Representation}

 \author{
 \Name{{Gerrit Gerhartz$^{*}$}$^1$, Peter Lippmann\thanks{equal contribution}$^1$  \& Fred A. Hamprecht$^1$ \phantom{{}}} \\
 \addr \textsuperscript{1}Interdisciplinary Center for Scientific Computing (IWR), Heidelberg University, Germany \\ 
 \Email{gerhartz@thphys.uni-heidelberg.de, peter.lippmann@iwr.uni-heidelberg.de}
 }

\jmlrauthors{G.G., P.L. \& F.A.H}

\makeatletter
\jmlrproceedings{PMLR}{Extended Abstract Track, NeurReps}


\makeatletter
\newcommand{\tweakjmlrmargins}[2]{%
  \setlength{\@tempdima}{#1}%
  \addtolength{\textwidth}{\@tempdima}%
  \setlength{\@tempdimb}{.5\@tempdima}%
  \addtolength{\oddsidemargin}{-\@tempdimb}%
  \addtolength{\evensidemargin}{-\@tempdimb}%

  \setlength{\@tempdima}{#2}%
  \addtolength{\textheight}{\@tempdima}%
  \addtolength{\topmargin}{-.5\@tempdima}%
  \addtolength{\headsep}{-.1\@tempdima}%
  \setlength{\headheight}{12pt}%
}
\makeatother

\makeatletter
\newcommand{\stretchpagelayout}[2]{%
  \setlength{\headheight}{12pt}%
  \addtolength{\textheight}{#1}%
  \addtolength{\textheight}{#2}%
  \addtolength{\topmargin}{-#2}%
}
\makeatother

\tweakjmlrmargins{0.10in}{0.10in} 
\stretchpagelayout{0.33in}{-0.13in}

\makeatletter
\jmlrpages{}   %
\makeatother

\begin{document}

\maketitle

\begin{abstract}
Equivariant neural networks offer strong inductive biases for learning from molecular and geometric data but often rely on specialized, computationally expensive tensor operations. We present a framework to transfers existing tensor field networks into the more efficient local canonicalization paradigm, preserving equivariance while significantly improving the runtime. Within this framework, we systematically compare different equivariant representations in terms of theoretical complexity, empirical runtime, and predictive accuracy. We publish the \texttt{tensor\_frames} package, a PyTorchGeometric based implementation for local canonicalization, that enables straightforward integration of equivariance into any standard message passing neural network.
\end{abstract}

\section{Introduction}
Molecular systems in 3D space exhibit fundamental spatial symmetries — for instance, the energy of a molecule remains invariant under global rotations and reflections, while vectorial properties such as dipole moments transform accordingly. Learning models that respect these symmetry constraints is essential for both accuracy and generalization. This has led to the development of equivariant neural networks, which enforce consistent transformation behavior of outputs with respect to the input geometry. 
However, two key challenges remain: (a) comparing equivariant models with data augmentation is non-trivial~\citep{lippmann2024tensor,brehmer2024does}, and (b) existing equivariant architectures often rely on specialized and computationally demanding building blocks~\citep{passaro2023reducing}.

A recent line of work addresses both issues through local canonicalization~\citep{lippmann2024tensor,spinner2025lorentz}, offering a lightweight and efficient way to enforce exact equivariance. In this work, we build upon and extend this framework to molecular machine learning: We show how to transfer existing equivariant tensor field networks into the framework of local canonicalization, achieving improved runtime at competitive accuracy and additional flexibility in the choice of possible representations. We systematically compare different equivariant representations in terms of theoretical complexity, runtime, and predictive performance. We release a modular, efficient PyTorchGeometric based implementation that enables easy integration of our formalism into any standard message passing network. Please find all code of our package at \url{https://github.com/sciai-lab/tensor_frames}.

\section{Background: group representations and equivariance}\label{sec:group_representations}
Symmetries in physical systems can be described using the mathematical foundations of group theory. Given a group $G$, a \textit{group representation} $\rho$ on a vector space $V$ is a group homomorphism $\rho: G \to GL(V)$ such that
\begin{equation} \label{eq:rep}
    \rho(g_1 g_2) = \rho(g_1) \rho(g_2) \quad \forall g_1, g_2 \in G,
\end{equation}
defining how elements $g \in G$ act on vectors $v \in V$, i.e.~$(\rho(g) v)_i = \sum_j \rho(g)_{ij} v_j$. 
A function $\varphi: V \to W$ is \textit{equivariant} under $G$ if $\rho_{\mathrm{out}}(g)\varphi(x)=\varphi(\rho_{\mathrm{in}}(g)x)$ for all $g \in G$ and $x\in V$, where $\rho_{\mathrm{in}}, \rho_{\mathrm{out}}$ are representations on $V$ and $W$ respectively. 

\paragraph{Representations of $\boldsymbol{\mathrm{O}(3)}$.}\label{sec:O3_representations}
We consider representations of the group $\mathrm{O}(3)$, the group of rotations and reflections in $\mathbb R^3$,  to describe how geometric data transforms in equivariant networks. A vector $v$ transforms under $R \in \mathrm{O}(3)$ as $(R v)_i = \sum_j R_{ij} v_j$. Correspondingly, higher-order \textit{Cartesian tensors} transform as
\begin{equation} \label{eq:tensor}
    T'_{i_1 \ldots i_n} = \sum_{j_1, ... , j_n} R_{i_1 j_1} \cdots R_{i_n j_n} T_{j_1 \ldots j_n} \,\, \text{ or } \,\, P'_{i_1 \ldots i_n} = \det(R) \sum_{j_1, ... , j_n} R_{i_1 j_1} \cdots R_{i_n j_n} P_{j_1 \ldots j_n}.
\end{equation}
We may distinguish between tensors $T$ and pseudotensors $P$ which behave differently under reflections, i.e.~orientation-reversing transformations~\citep{jeevanjee2011introduction}. The representations in Eq.~\eqref{eq:tensor} can be decomposed into so-called \textit{irreducible representations} (see App.~\ref{sec:decomposition}). The irreducible representations of $\mathrm{SO}(3)$ are indexed by $l \in \mathbb{N}_0$ and described by the Wigner-$D$ matrices $D^{(l)}(R)$, which act on $(2l + 1)$-dimensional tensors $x$ as $(D^{(l)}(R) x)_m = \sum_{m'} D^{(l)}_{mm'}(R) x_{m'}$.
Internal representations in neural networks often combine multiple geometric types into a direct sum of representations. 

\paragraph{Equivariance via local canonicalization.}

The key idea of equivariance by local canonicalization~\citep{lippmann2024tensor} is the following: Based on the Euclidean geometry of the network input, one predicts one equivariant local frame $R_i$ at each node $i$. The geometric input node features $F_i$ are transformed from the global frame of reference into the local frames, yielding coordinates $f_i = \rho_{in}(R_i) F_i$ invariant to the choice of global frame. After this canonicalization step, the node features can be processed using an arbitrary backbone architecture without breaking the invariance.
However, in order to communicate geometric information during message passing between nodes with distinct local frames, it is crucial that \textit{tensorial} messages are transformed from one local frame into the other. This yields the following general form of invariant message passing \textit{with tensorial messages}, as proposed in~\citep{lippmann2024tensor}:
\begin{equation}\label{eq:tensor_msg}
      f_i^{(k)}=\bigoplus_{j\in\mathcal{N}(i)}\phi^{(k)}\Big(\rho_{\mathrm f}(R_i R_j^{-1})f_j^{(k-1)}, R_i( x_i - x_j)\Big) \bigg) .
\end{equation}
The internal message representation $\rho_{\mathrm f}$ can be chosen freely as a hyperparameter.

\section{Learning representations}\label{sec:learning_representations}
In many problems, there is no canonical choice for the representation under which geometric messages should transform in Eq.~\eqref{eq:tensor_msg}. It can therefore be beneficial for the model to learn the transformation from frame $R_j$ to $R_i$. However, enforcing the strict mathematical properties of a group representation, cf.~Eq.~\eqref{eq:rep}, for a learned transformation is challenging. A more flexible alternative is to relax this requirement and not enforce Eq.~\eqref{eq:rep}. As long as the transition matrix $R_iR_j^{-1}$ is accounted for in the message passing, geometric consistency can in principle be preserved (cf.~Fig.~1 in~\citep{lippmann2024tensor}). The simplest approach would be to learn a linear transformation between local frames. However, this requires outputting a full matrix of size $d_f \times d_f$ from an MLP, which scales poorly with feature dimension $d_f$. To address this, we propose learning the effect of the frame transition on the features directly: $\rho_f(R_i R_j^{-1}) f_j \ \Rightarrow \ \mathtt{MLP}(R_i R_j^{-1}, f_j)$.
This learned transformation allows the model to flexibly adapt the transformation behavior to the task, eliminating the need for manual tuning.

\section{Transforming tensor field based architectures to local canonicalization} \label{sec:modern_architectures}
The typical tensor field network is based on internal features that transform under the irreducible representation and messages are computed via a tensor product convolution:
\begin{align}\label{eq:tensor_product_message}
    m_{ij,m_3}^{(l_3)}=\left[f_j^{(l_1)}\otimes \mathcal R(r_{ij})Y^{(l_2)}(\hat{r}_{ij})\right]_{m_3} = \sum_{m_1=-l_1}^{l_1}\sum_{m_2=-l_2}^{l_2}C_{m_1l_1,m_2l_2}^{m_3l_3} f^{(l_1)}_{j,m_1} \mathcal R(r_{ij})Y^{(l_2)}_{m_2}(\hat{r}_{ij}),
\end{align}
where $\mathcal R(r_{ij})$ is a radial embedding of $r_{ij} = \| x_i - x_j \|$, $Y^{(l_2)}(\hat{r}_{ij})$ are spherical harmonics evaluated on the unit vector $\hat r_{ij}=\frac{x_i - x_j}{\| x_i - x_j \|}$, and $C$ are Clebsch–Gordan coefficients that couple irreducible representations of the angular momenta $l_1$, $l_2$, and $l_3$. This operation combines three ingredients: node features, a radial function of distance, and an angular component via spherical harmonics. Motivated by this, we design a neural network layer that mimics this structure using standard operations:
\begin{align}\label{eq:edge_linear_layer_with_transform}
    \mathtt{EDGE}(\rho_{\mathrm{f}}(R_iR_j^\mathrm{T})f_j,R_i (x_i - x_j))=A(B\rho_{\mathrm{f}}(R_iR_j^{-1})f_j\odot \texttt{MLP}(\mathcal R(r_{ij}) ||\Theta(R_i\hat{r}_{ij}))) .
\end{align}
\begin{wrapfigure}{r}{0.33\textwidth}
\vspace{-0.3cm}
 \centering
    \includegraphics[width=0.85\linewidth]{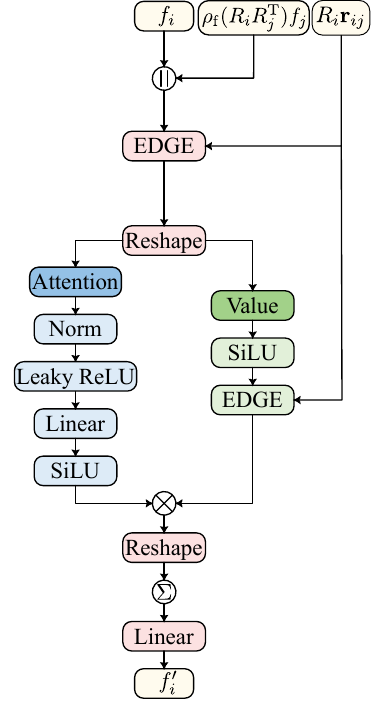}
    \vspace{-0.4cm}
    \caption{\textbf{LoCaFormer attention block.}}
    \label{fig:attention_mechanisms}
    \vspace{-.6cm}
\end{wrapfigure}
Following the form of Eq.~\ref{eq:tensor_msg}, here, $\mathcal R$ and $\Theta$ are radial and angular embedding functions of the local relative distance vector $R_i(x_i - x_j)$, see App.~\ref{app:rad_ang} and $A$ and $B$ are learned linear transformations. $\odot$ denotes element-wise multiplication and $||$ indicates concatenation. The transformation $\rho_{\mathrm{f}}(R_i R_j^\mathrm{T})$ transforms features from node $j$ into the local frame of node $i$. This EDGE layer serves as a drop-in replacement for the tensor product in Eq.~\eqref{eq:tensor_product_message}, enabling expressive equivariant message passing while relying on efficient, standard components.

Using local canonicalization and the \texttt{EDGE} building block, we have implemented an attention-based message passing architecture, the LoCaFormer~(Fig.~\ref{fig:attention_mechanisms}), inspired by the widely used Equiformer architecture~\citep{liao2022equiformer}, see App.~\ref{app:ex_details} for details. Beyond an implementation of the LoCaFormer layer, our python package \texttt{tensor\_frames} includes a module to predict equivariant local frames as proposed in~\citep{lippmann2024tensor}, efficient implementations for the transformation of the irreducible representation, Cartesian tensor representation and the MLP representation described in Sec.~\ref{sec:learning_representations}. The main module of the package is \texttt{tensor\_frames.nn.TFMessagePassing} which can be used as a drop-in replacement for the widely used message passing module in PytorchGeometric~\citep{Fey/Lenssen/2019} to make any existing message passing layer equivariant using local canonicalization combined with tensorial message passing, see App.~\ref{sec:software}.
\vspace{-0.1cm}

\section{Discussion and results on molecular property prediction}\label{sec:modern_local_canonicalization}

We have trained four LoCaFormer models with different internal representations on the QM9 dataset~\citep{wu2018moleculenet}, see Tab.~\ref{tab:qm9}.
The models achieve competitive performance, while being between 4-and 5-times faster than the widely used Equiformer, see App.~\ref{app:ex_details} for experimental details. Tensorial messages with Cartesian tensors or irreducible representations improve predictions for the isotropic polarizability $\alpha$ and magnetic dipole moment norm $\mu$, likely because these properties depend more strongly on geometric information. To test this, we have  trained the same models on the dataset of~\citet{zhang2023learning}, containing 10,000 N-methylacetamide configurations with dipole moments $\mu$ (vectors) and polarizabilities $\alpha$ (rank-2 tensors). Here, tensorial message passing notably outperforms both scalar message passing and the learned MLP representation, confirming its advantage for tensorial targets. This experiment highlights the flexibility of our framework, which allows straightforward comparison across representations to decide whether the extra cost of tensorial features is justified or invariant features suffice. Notably, the less common Cartesian tensor representation can be computationally more efficient than irreducible representations when the transformation of the latter is limited by the Wigner matrix computation, see App.~\ref{app:complexity}.

\begin{table}[ht]
\centering
\vspace{-0.4cm}
\begin{minipage}{0.45\linewidth}
\centering
\caption{\textbf{MAE and runtime on QM9 property prediction.}}
\scalebox{0.65}{
  \begin{tabular}{@{}lccccccc@{}}
    \toprule
    Method & $\underset{[a_0]}{\alpha}$ &  $\underset{[\text{meV}]}{\epsilon_{\text{HOMO}}} $  &  $\underset{[\text{meV}]}{\epsilon_{\text{LUMO}}}$ & $\underset{[\text{D}]}{\mu}$ & $\underset{[\mathrm s^-1]}{\text{it/s}}$ \\
    \midrule
    Equiformer \citep{liao2022equiformer} & \underline{.050} & \textbf{14} & \textbf{13} & \textbf{.010} & 0.8 \\
    MACE \citep{batatia2022mace} & \textbf{.038} & 22 & 19 & \underline{.015} & n/a\\
    \midrule
    LoCaFormer: Scalar messages & .057 & 20.8 & 18.2 & .030 & 4.5\\
    LoCaFormer: Cart. tensor rep. & .052 &  20.6 &  \underline{17.7} & .018 & 3.8\\
    LoCaFormer: Irreducible rep. & .054 &  \underline{19.1} & 19.4 & .020 & 3.3\\
    LoCaFormer: MLP rep. & .057 & 20.6 & 17.9 & .030 & 4.0\\
    \bottomrule
  \end{tabular}}
  \label{tab:qm9}
\end{minipage}%
\hfill
\begin{minipage}{0.45\linewidth}
\centering
\caption{\textbf{RMSE of magn. dipole moment $\mu$ and polarizability $\alpha$.}}
\scalebox{0.7}{
  \begin{tabular}{@{}lcccccccc@{}}
    \toprule
    Method & $\mu \,[\text{a.u.}]$ & $\alpha\, [\text{a.u.}] $ \\
    \midrule
    EANN \citep{zhang2023learning, zhang2020efficient} & .004 & .020 \\
    \midrule
    LoCaFormer: Scalar messages & .005 & .050 \\
    LoCaFormer:  Cart. tensor rep. & .003 & .036 \\
    LoCaFormer:  Irreducible rep. & .003 & .042 \\
    LoCaFormer:  MLP rep. & .004 & .158 \\
    \bottomrule
  \end{tabular}}
  \label{tab:tensorial_chemical_properties}
\end{minipage}
\vspace{-0.1cm}
\end{table}

Furthermore, we have compared our best performing LoCaFormer models (according to Tab.~\ref{tab:qm9}) against models trained without local canonicalization but using data augmentation instead.
In our experiments (App.~\ref{app:relation_data_augmentation}), the models with built-in equivariance are more data efficient, as expected, meaning that the prediction error decreases faster as more training data becomes available. However, perhaps surprisingly, in the low data regime the prediction error of the data augmented model is smaller. This challenges the common belief that models with built-in equivariance should prevail in the low data regime since they do not need any extra data to model the symmetries; and aligns with results in~\citep{lippmann2024tensor} for a completely different architecture and dataset.

To summarize, the framework of local canonicalization offers a promising alternative for equivariant architectures in molecular ML. Our comparative study highlights trade-offs between different internal representations. While learned representations remain less effective than exact group representations, we believe that learning representations can be an interesting avenue for future research with numerous design choices yet to be explored. By open-sourcing our efficient implementation we would like to aid the development of the field of local canonicalization.

\acks{This work is supported by the Klaus Tschira Stiftung gGmbH (SIMPLAIX project P4) and by Deutsche Forschungsgemeinschaft (DFG) under Germany’s Excellence Strategy EXC-2181/1 - 390900948 (the Heidelberg STRUCTURES Excellence Cluster) as well as under project number 240245660 - SFB 1129.}

\appendix

\bibliography{references}

\begin{thebibliography}{17}
\providecommand{\natexlab}[1]{#1}
\providecommand{\url}[1]{\texttt{#1}}
\expandafter\ifx\csname urlstyle\endcsname\relax
  \providecommand{\doi}[1]{doi: #1}\else
  \providecommand{\doi}{doi: \begingroup \urlstyle{rm}\Url}\fi

\bibitem[Batatia et~al.(2022)Batatia, Kovacs, Simm, Ortner, and
  Cs{\'a}nyi]{batatia2022mace}
Ilyes Batatia, David~P Kovacs, Gregor Simm, Christoph Ortner, and G{\'a}bor
  Cs{\'a}nyi.
\newblock {MACE}: Higher order equivariant message passing neural networks for
  fast and accurate force fields.
\newblock \emph{Advances in Neural Information Processing Systems},
  35:\penalty0 11423--11436, 2022.

\bibitem[Batzner et~al.(2022)Batzner, Musaelian, Sun, Geiger, Mailoa,
  Kornbluth, Molinari, Smidt, and Kozinsky]{batzner20223}
Simon Batzner, Albert Musaelian, Lixin Sun, Mario Geiger, Jonathan~P Mailoa,
  Mordechai Kornbluth, Nicola Molinari, Tess~E Smidt, and Boris Kozinsky.
\newblock E(3)-equivariant graph neural networks for data-efficient and
  accurate interatomic potentials.
\newblock \emph{Nature communications}, 13\penalty0 (1):\penalty0 2453, 2022.

\bibitem[Brehmer et~al.(2024)Brehmer, Behrends, De~Haan, and
  Cohen]{brehmer2024does}
Johann Brehmer, S{\"o}nke Behrends, Pim De~Haan, and Taco Cohen.
\newblock Does equivariance matter at scale?
\newblock \emph{arXiv preprint arXiv:2410.23179}, 2024.

\bibitem[Ebert et~al.(2003)Ebert, Musgrave, Peachey, Perlin, and
  Worley]{ebert2003texturing}
David~S Ebert, F~Kenton Musgrave, Darwyn Peachey, Ken Perlin, and Steven
  Worley.
\newblock Texturing and modeling: A procedural approach, with contributions
  from wr mark and jc hart, 2003.

\bibitem[Fey and Lenssen(2019)]{Fey/Lenssen/2019}
Matthias Fey and Jan~E. Lenssen.
\newblock Fast graph representation learning with {PyTorch Geometric}.
\newblock In \emph{ICLR Workshop on Representation Learning on Graphs and
  Manifolds}, 2019.

\bibitem[Gasteiger et~al.(2020)Gasteiger, Groß, and
  Günnemann]{Gasteiger2020Directional}
Johannes Gasteiger, Janek Groß, and Stephan Günnemann.
\newblock Directional message passing for molecular graphs.
\newblock In \emph{International Conference on Learning Representations}, 2020.
\newblock URL \url{https://openreview.net/forum?id=B1eWbxStPH}.

\bibitem[Hestness et~al.(2017)Hestness, Narang, Ardalani, Diamos, Jun,
  Kianinejad, Patwary, Yang, and Zhou]{hestness2017deep}
Joel Hestness, Sharan Narang, Newsha Ardalani, Gregory Diamos, Heewoo Jun,
  Hassan Kianinejad, Md~Mostofa~Ali Patwary, Yang Yang, and Yanqi Zhou.
\newblock Deep learning scaling is predictable, empirically.
\newblock \emph{arXiv preprint arXiv:1712.00409}, 2017.

\bibitem[Jeevanjee(2011)]{jeevanjee2011introduction}
Nadir Jeevanjee.
\newblock \emph{An introduction to tensors and group theory for physicists}.
\newblock Springer, 2011.

\bibitem[Liao and Smidt(2022)]{liao2022equiformer}
Yi-Lun Liao and Tess Smidt.
\newblock Equiformer: Equivariant graph attention transformer for 3d atomistic
  graphs.
\newblock \emph{arXiv preprint arXiv:2206.11990}, 2022.

\bibitem[Lippmann et~al.(2025)Lippmann, Gerhartz, Remme, and
  Hamprecht]{lippmann2024tensor}
Peter Lippmann, Gerrit Gerhartz, Roman Remme, and Fred~A Hamprecht.
\newblock Beyond canonicalization: How tensorial messages improve equivariant
  message passing.
\newblock In Y.~Yue, A.~Garg, N.~Peng, F.~Sha, and R.~Yu, editors,
  \emph{International Conference on Representation Learning}, volume 2025,
  pages 88067--88087, 2025.
\newblock URL
  \url{https://proceedings.iclr.cc/paper_files/paper/2025/file/db7534a06ace69f4ec95bc89e91d5dbb-Paper-Conference.pdf}.

\bibitem[Passaro and Zitnick(2023)]{passaro2023reducing}
Saro Passaro and C~Lawrence Zitnick.
\newblock Reducing {SO(3)} convolutions to {SO(2)} for efficient equivariant
  gnns.
\newblock In \emph{International Conference on Machine Learning}, pages
  27420--27438. PMLR, 2023.

\bibitem[Pinchon and Hoggan(2007)]{pinchon2007rotation}
Didier Pinchon and Philip~E Hoggan.
\newblock Rotation matrices for real spherical harmonics: general rotations of
  atomic orbitals in space-fixed axes.
\newblock \emph{Journal of Physics A: Mathematical and Theoretical},
  40\penalty0 (7):\penalty0 1597, 2007.

\bibitem[Spinner et~al.(2025)Spinner, Favaro, Lippmann, Pitz, Gerhartz, Plehn,
  and Hamprecht]{spinner2025lorentz}
Jonas Spinner, Luigi Favaro, Peter Lippmann, Sebastian Pitz, Gerrit Gerhartz,
  Tilman Plehn, and Fred~A Hamprecht.
\newblock Lorentz local canonicalization: How to make any network
  lorentz-equivariant.
\newblock \emph{arXiv preprint arXiv:2505.20280}, 2025.

\bibitem[Wang et~al.(2019)Wang, Sun, Liu, Sarma, Bronstein, and
  Solomon]{wang2019dynamic}
Yue Wang, Yongbin Sun, Ziwei Liu, Sanjay~E Sarma, Michael~M Bronstein, and
  Justin~M Solomon.
\newblock Dynamic graph cnn for learning on point clouds.
\newblock \emph{ACM Transactions on Graphics (tog)}, 38\penalty0 (5):\penalty0
  1--12, 2019.

\bibitem[Wu et~al.(2018)Wu, Ramsundar, Feinberg, Gomes, Geniesse, Pappu,
  Leswing, and Pande]{wu2018moleculenet}
Zhenqin Wu, Bharath Ramsundar, Evan~N Feinberg, Joseph Gomes, Caleb Geniesse,
  Aneesh~S Pappu, Karl Leswing, and Vijay Pande.
\newblock Moleculenet: a benchmark for molecular machine learning.
\newblock \emph{Chemical science}, 9\penalty0 (2):\penalty0 513--530, 2018.

\bibitem[Zhang et~al.(2020)Zhang, Ye, Zhang, Hu, Jiang, and
  Jiang]{zhang2020efficient}
Yaolong Zhang, Sheng Ye, Jinxiao Zhang, Ce~Hu, Jun Jiang, and Bin Jiang.
\newblock Efficient and accurate simulations of vibrational and electronic
  spectra with symmetry-preserving neural network models for tensorial
  properties.
\newblock \emph{The Journal of Physical Chemistry B}, 124\penalty0
  (33):\penalty0 7284--7290, 2020.

\bibitem[Zhang et~al.(2023)Zhang, Jiang, and Jiang]{zhang2023learning}
Yaolong Zhang, Jun Jiang, and Bin Jiang.
\newblock Learning dipole moments and polarizabilities.
\newblock In \emph{Quantum Chemistry in the Age of Machine Learning}, pages
  453--465. Elsevier, 2023.

\end{thebibliography}

\newpage

\appendix

\section{Software contribution}\label{sec:software}
We have implemented the general local canonicalization framework of~\citet{lippmann2024tensor} as a modular, publicly available Python library. All components, such as learning local coordinate frames and handling different representations, are provided as reusable \texttt{PyTorch} modules, enabling direct integration into existing architectures.

\texttt{PyTorchGeometric}~\citep{Fey/Lenssen/2019} is one of the most commonly used Python libraries to implement graph neural networks. For concreteness, let us consider a standard message passing network with message passing layers of the following form
\begin{align}
    f_i^{(k+1)} = \bigoplus_{j \in \mathcal N(i)}\phi(f_j^{(k)}, x_i - x_j),
\end{align}
where $f_i$ are the local node features at node $i$ and $x_i - x_j$ is the relative distance vector. 
Using local canonicalization with local frames $R_i$ at each node $i$ and tensorial messages, the modified message passing formula reads
\begin{equation}
    f_i^{(k+1)} = \bigoplus_{j \in \mathcal N(i)}\phi(\rho_{\mathrm f}(R_i R_j^{-1})f_j^{(k)}, R_i (x_i - x_j)).
\end{equation}
The above can be implemented effortlessly in our \texttt{tensor\_frames} packages whose main module \texttt{TFMessagePassing} is a wrapper around \texttt{torch\_geometric.nn.MessagePassing} that handles all transformations automatically after providing the used representations. The following minimal code example in Listing~\ref{lst:edge_conv_pyg} and~\ref{lst:edge_conv_tf}  illustrates how easily one can transfer existing non-equivariant message passing layers to layers which exhibit exact built-in equivariance using the framework of local canonicalization. Notably, the message function, which in this example is very simple but often contains the most complicated logic of the layer, must not be altered at all. Our full Python code is available at \url{https://github.com/sciai-lab/tensor_frames}.

\begin{lstlisting}[style=mypython, caption={Example of a variant of the \texttt{EdgeConv} layer~\citep{wang2019dynamic} implemented in \texttt{PyTorchGeometric}.}, label={lst:edge_conv_pyg}]
    from torch_geometric.nn import MessagePassing as MP
    from torch_geometric.models import MLP
    
    class EdgeConv(MP):
        def __init__(self, in_dim, out_dim):
            super().__init__()
            self.mlp = MLP(in_dim + 3, out_dim, hidden_channels=[256])
    
        def forward(self, edge_index, f, pos):
            return self.propagate(edge_index, f=f, pos=pos)
    
        def message(self, f_j, pos_i, pos_j):
            return self.mlp(torch.cat([f_j, pos_i - pos_j], dim=-1))
    
    message_passing_layer = EdgeConv(32, 32)
\end{lstlisting}
\newpage

\begin{lstlisting}[style=mypython, caption={Example of equivariant adaptation of the \texttt{EdgeConv} layer variant using our \texttt{TFMessagePassing} for tensorial message passing in local canonicalization. The module is a wrapper around \texttt{torch\_geometric.nn.MessagePassing}.}, label={lst:edge_conv_tf}]
@r-    from torch_geometric.nn import MessagePassing @
@g+    from tensor_frames.nn.tfmessage_passing import TFMessagePassing as MP @
@g+    from tensor_frames.reps Irreps, MLPReps, TensorReps @
@g+    from tensor_frames.lframes.lframes import LFrames @
    from torch_geometric.models import MLP
    
    class EdgeConv(MP):
@r-        def __init__(self, in_dim, out_dim): @
@g+        def __init__(self, in_reps: Irreps | TensorReps | MLPReps, out_reps): @
@r-            super().__init__() @
@g+            super().__init__( @
@g                params_dict={ @
@g                    "x": {"type": "local", "rep": in_reps} @
@g                    "positions": {"type": "global", "rep": TensorReps("1x0n")}, @
@g                } @
@g            ) @
@r-            self.mlp = MLP(in_dim + 3, out_dim, hidden_channels=[256]) @
@g+            self.mlp = MLP(in_reps.dim + 3, out_reps.dim, hidden_channels=[256]) @
    
@r-        def forward(self, edge_index, f, pos): @
@g+        def forward(self, edge_index, f, pos, lframes: LFrames): @
@r-            return self.propagate(edge_index, f=f, pos=pos) @
@g+            return self.propagate(edge_index, f=f, pos=pos, lframes=lframes) @
    
        def message(self, f_j, pos_i, pos_j):
            return self.mlp(torch.cat([f_j, pos_i - pos_j], dim=-1))
    
@r-    message_passing_layer = EdgeConv(32, 32) @
@g+    message_passing_layer = EdgeConv(Irreps("8x0n+8x1n"), Irreps("1x0n+1x1n")) @
\end{lstlisting}

\section{Connection between irreducible and Cartesian tensor representation}\label{sec:decomposition}
In this section, we demonstrate how the Cartesian tensor representations used in our experiments can be decomposed into irreducible representations to allow for a fair comparison.

A representation is called \textit{irreducible} if there is no non-trivial subspace of the vector space that the representation acts on that is closed under the action of the group \citep{jeevanjee2011introduction}. In other words, the irreducible representations can be thought of as the smallest building blocks of a representation, which cannot be further decomposed into smaller representations.

The irreducible representations of $\mathrm{SO}(3)$ are given by the Wigner-$D$ matrices, which act on $2l+1$-dimensional vector spaces. Therefore, $l$ is an index that labels the irreducible representations. An element of the vector space is a vector with $2l+1$ components, which is called a spherical tensor. For a spherical tensor $x$, the Wigner-$D$ matrix acts as follows:
\begin{align} \label{eq:wigner_trafo}
    (D^{(l)}(R) x)_m = \sum_{m'=-l}^l D^{(l)}_{mm'}(R) x_{m'}, \quad m\in \{-l,\dots, l\}.
\end{align}
Inside equivariant neural networks, it is common to use features that combine several different representations, such as vectors and scalars. 

Let us illustrate for a rank-$2$ tensor representation $T_{ij}$ of $\mathrm{SO}(3)$ the decomposition into irreducible representations. 
First the symmetric $S_{ij}$ and the antisymmetric $A_{ij}$ parts of the rank-$2$ tensor $T_{ij}$
\begin{align}
    S_{ij} = \frac{1}{2}(T_{ij}+T_{ji}), \quad A_{ij} = \frac{1}{2}(T_{ij}-T_{ji}),
\end{align}
are defined, which form representations themselves. The symmetric part lives in a subspace closed under the action of the group, which can be seen by
\begin{align}
    S'_{ij} = \frac{1}{2}(T'_{ij}+T'_{ji}) = \frac{1}{2}(R_{ik}R_{jl}T_{kl}+R_{jk}R_{il}T_{kl}) = R_{ik}R_{jl}\frac{1}{2}(T_{kl}+T_{lk}) = R_{ik}R_{jl}S_{kl}.
\end{align}
This proof also holds for the antisymmetric part. Counting degrees of freedom, antisymmetric part is a three-dimensional subspace, which can be shown to be equivalent to the irreducible representation of $l=1$. The symmetric part can be further decomposed into a trace and the traceless part. Both forming its own subspace:
\begin{align}
    \text{Tr}(S') = \sum_i S'_{ii} = \sum_{j,k}\sum_i R_{ij}R_{ik}S_{jk} = \sum_i S_{ii} = \text{Tr}(S),
\end{align}
where we have used that $RR^\top=1$. The trace is the subspace, which corresponds to the irreducible representation of $l=0$. The last part, which is left, is the traceless symmetric part
\begin{align}
    S_{ij}-\delta_{ij}\frac{\text{Tr}(S)}{3},
\end{align}
which is an irreducible subspace and corresponds to the irreducible representation of $l=2$ with dimension 5. 
Note that $T_{ij}$ transforms like the outer product of two vectors $v$ and $w$, e.g. $T'_{ij} = v'_i w'_j = R_{ik} R_{jl} v_k w_l = R_{ik} R_{jl} T_{kl}$. This justifies that the Cartesian tensors of rank 2 is labeled by its dimensions $\underline{3} \otimes \underline{3}$, where $\underline{3}$ corresponds to the irreducible representation of $l=1$ that has dimension 3. In terms of dimensions the decomposition of a rank-2 Cartesian tensor into irreducible representations is given by:
\begin{align}
    \underline{3}\otimes \underline{3}&=\underline{1}\oplus \underline{3} \oplus \underline{5}.
\end{align}
The general rule to decompose the outer product or tensor product of two spherical tensors with angular momenta $l_1$ and $l_2$ is:
\begin{align}
    \underline{2l_1 + 1} \otimes \underline{2l_2 + 1} = \bigoplus_{L=|l_1 - l_2|}^{l_1 + l_2} \underline{2L + 1}
\end{align}
Following this rule, we can decompose higher-order Cartesian tensor representations:
\begin{align}
    \underline{3}\otimes \underline{3}\otimes \underline{3}&=(\underline{1}\oplus \underline{3} \oplus \underline{5})\otimes \underline{3} \nonumber \\
    &=\underline{1} \otimes \underline{3} \oplus \underline{3} \otimes \underline{3} \oplus \underline{5} \otimes \underline{3} \nonumber\\
    &= \underline{3} \oplus (1\oplus \underline{3} \oplus \underline{5})  \oplus (\underline{3}\oplus \underline{5} \oplus \underline{7}) \nonumber \\
    &= \underline{1} \oplus \underline{3} \oplus \underline{3} \oplus \underline{3} \oplus \underline{5} \oplus \underline{5} \oplus \underline{7} .
\end{align}
Using these decompositions, the Cartesian tensor representations used in the experiments can be decomposed into irreducible representations to allow for a fair comparison.

\section{Computational complexity of Cartesian tensors and irreducible representations} \label{app:complexity}

\begin{table}[t]
    \caption{\textbf{Computational complexity for Cartension tensor representation and irreps of $\boldsymbol{\mathrm{SO}(3)}$} ``Per entry'' refers to the cost of transforming a single feature component. The total transform cost for the irreducible representation includes the computation of the Wigner-D matrix, which is the complexity bottleneck.}
    \vspace{5pt}
    \centering
    \begin{tabular}{@{}lccc@{}}
        \toprule
        Representation & \# Components & Total Transform Cost & Per-Entry Cost \\
        \midrule
        Irrep ($l$) & $2l+1$ & $\mathcal{O}(l^3)$ & $\mathcal{O}(l^2)$ \\
        Cart. tensor rep ($n$) & $3^n$ & $\mathcal{O}(n \, 3^n)$ & $\mathcal{O}(n)$ \\
        \bottomrule
    \end{tabular}
    \label{tab:reps_comparison}
\end{table}

Let $n$ denote the order of a Cartesian tensor representation in $3$-dimensional Euclidean space. The highest irreducible representation contained in such a tensor is $l=n$, as discussed in App.~\ref{sec:decomposition}. A tensor of order $n$ has $3^n$ components, and the transformation of a single component under a group element $R \in \mathrm{O}(3)$ is given by
\begin{align}
    T'_{i_1 \dots i_n} = \sum_{j_1, ..., j_n} R_{i_1 j_1} \dots R_{i_n j_n} T_{j_1 \dots j_n},
\end{align}
involving $n$ contracted indices, each running over $d=3$ values. Updating one component therefore requires $\mathcal{O}(n)$ operations, and the full tensor transformation scales as $\mathcal{O}(n \, 3^n)$.

For comparison, consider a 3D irreducible representation of angular momentum $l$. Such a representation has $2l+1$ components, which transform according to the Wigner-$D$ matrix $D^{(l)} \in \mathbb{R}^{(2l+1) \times (2l+1)}$, cf.~Eq.~\eqref{eq:wigner_trafo}.
Applying $D^{(l)}$ to a single component requires $\mathcal{O}(2l+1)$ operations. For the irreducible representation, the number of components grows linearly in $l$, while for the Cartesian tensor representation the number of components are $3^n$ for $n=l$. While Cartesian tensor representations grow much faster in components, they avoid the need to compute Wigner-$D$ matrices, whose computation is non-trivial.

The most efficient approach for computing the Wigner-$D$ matrices for a given $R \in \mathrm{SO}(3)$ is described in~\citep{pinchon2007rotation}, in which one uses different precomputed building blocks for every $l$. The approach scales like $\mathcal{O}(l^3)$ for the computation of $D^{(l)}(R)$. In practice, only one Wigner matrices must be computed for each combination of $l$ and $R$, so that it can be reused across feature channels of the same $l$. Nonetheless, for large $l$ the $\mathcal{O}(l^3)$ scaling may become a bottleneck. Therefore, if each feature component carries comparable information content, Cartesian tensor representations can be computationally more efficient when the computation of the Wigner matrix dominates the runtime. However, the need for large $l$ in practical applications is not yet fully explored, most likely related to this computational bottleneck.

\section{Radial and angular embedding of the molecule geometry} \label{app:rad_ang}
\begin{figure}[t]
    \centering
    \includegraphics[width=0.6\linewidth]{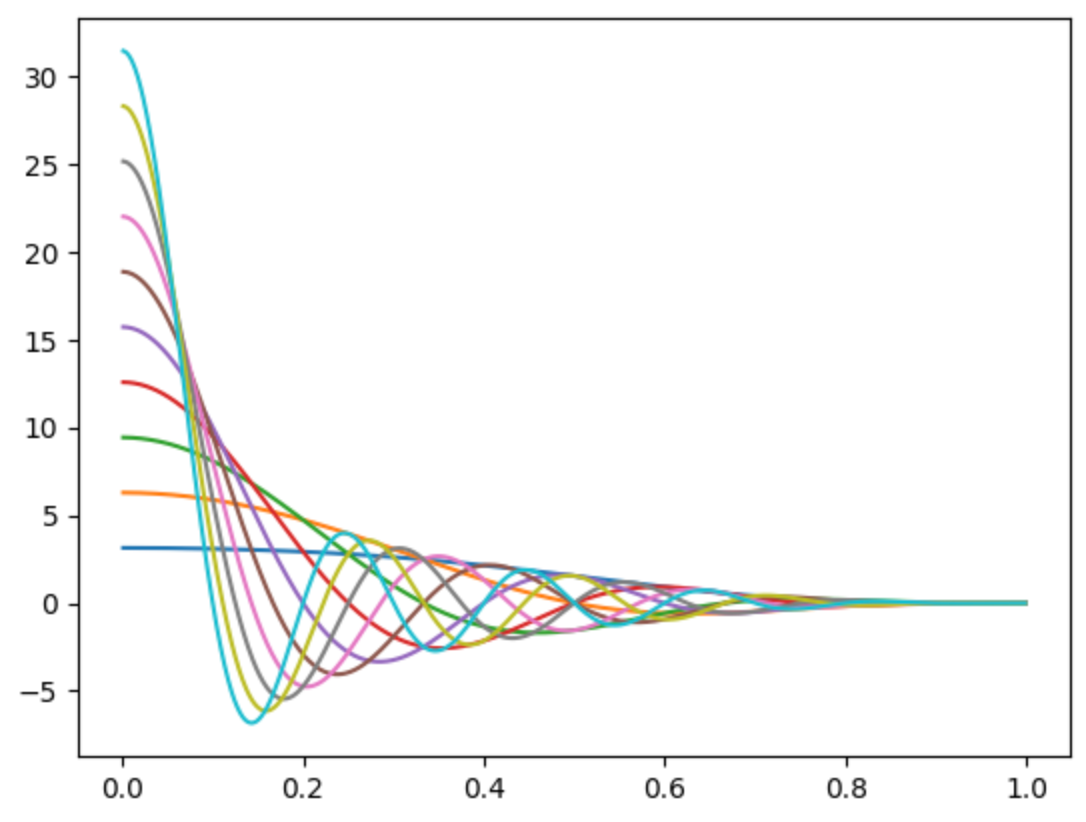}
    \caption{\textbf{Bessel embedding functions.} Here, Bessel functions of the first kind are depicted with ten different frequencies. The Bessel functions are also multiplied by the envelope function with a cutoff radius of $1$.}
    \label{fig:bessel}
\end{figure}

 The radial embedding of the relative distance $r_{ij} = \| x_i - x_j \|$ is calculated from Bessel functions of the first kind:
\begin{align}
    \mathcal R^{(m)}(r_{ij})=\frac{\omega(r_{ij})}{r_{ij}}\sin\left({\frac{r_{ij}}{r_c}\lambda^{(m)}}\right)
\end{align}
Here, $\omega$ is a smooth cutoff function~\citep{ebert2003texturing} that goes smoothly to $0$ at the cutoff radius $r_{c}$ and $\lambda^{(m)}$ are learnable frequencies. The embedding functions are depicted in Fig.~\ref{fig:bessel}. In principle, one could use other functions as the embedding functions such as Gaussians. We ablated this choice in preliminary experiments, and found Bessel functions performed equally well with fewer frequencies than Gaussian embedding functions. This coincides with the results of~\citet{Gasteiger2020Directional}.
We embed the angular part of the relative distance vector similarly to the radial part. Each component of the normalized relative distance vector $\hat{r}_{ij} = \frac{x_i - x_j}{\| x_i - x_j\|}$ is embedded separately:
\begin{equation}
    \theta^{(m)}(\hat{r}^k_{ij})=
    \frac{\omega(\hat{r}_{ij}^k)}{\hat{r}_{ij}^k}\sin\left({\hat{r}_{ij}^k\lambda^{(m)}}\right)
    \text{sign}({\hat{r}_{ij}^k}),\quad k\in \{x,y,z\},\quad \Theta=\theta(\hat{r}_{ij}^x) \ || \ \theta (\hat{r}_{ij}^y) \ ||\ \theta (\hat{r}_{ij}^z)
\end{equation}
Each component $k\in \{x,y,z\}$ of the normalized relative distance vector is embedded by  Bessel functions of the first kind, as for the radial embedding. This embedding needs to be multiplied with the sign of the component of the relative distance vector, since the Bessel functions are symmetric around the origin, but in the angular case negative and positive $x,y \text{ or } z$ values need to be embedded separately. Including the sign functions introduces a non-continuity if $x=0$, $y=0$ or $z=0$ in the angular embedding, but in experiments we not found found this to be a problem for the training dynamics of the model. The full angular embedding $\theta_{ij}$ is the concatenation of the three components.

\section{Experimental details}\label{app:ex_details}

\paragraph{Hyperparameter choices.} The hyperparameters for the models trained on QM9 property prediction are summarized in Tab.~\ref{tab:hyperparameters_chem}. Further, let us introduce the following notation to specify the feature representation used during message passing:  The feature representation is given by a direct sum of Cartesian tensor and pseudotensor representations. A Cartesian tensor representation is characterized by its order (i.e.~the number of indices, see~Eq.~\eqref{eq:tensor}) and its behavior under parity (\texttt{n} for tensors and \texttt{p} for pseudotensors). Furthermore, we specify the multiplicities, that is, how often each representation appears in a direct sum representation. For instance, the representation denoted as \texttt{8x0p+4x1n} is the direct sum of 8 pseudoscalars and 4 vectors. For a fair representation comparison, the decomposition of the Cartesian tensor representation into irreducible representations (cf.~App.~\ref{sec:decomposition}) can be used to build the corresponding irreducible representation. For the MLP representation the same feature dimension as in the other representations is used.

\begin{table}[t]
    \caption{\textbf{Hyperparameters for training our LoCaFormer models on QM9 property prediction.}}
  \centering
  \begin{tabular}{@{}lccc@{}}
    \toprule
    & LoCaFormer \\
    \midrule
    optimizer & AdamW \\
    weight decay & 5e-3 \\
    learning rate & 5e-4 \\
    scheduler & Cosine-LR \\
    epochs & 600 \\
    warm up epochs & 5 \\
    gradient clip & 0.5\\
    loss & Smooth-L1 \\
    \bottomrule
  \end{tabular}
  \label{tab:hyperparameters_chem}
\end{table}

\begin{table}[t]
    \caption{\textbf{Architecture of the LoCaFormer model.}}
    \vspace{0.4cm}
    \centering
    \begin{tabular}{@{}cc@{}}
        \toprule
        Parameter & Value \\
        \midrule
        Number of radial Bessel functions & 32 \\
        Number of angular Bessel functions & 20 \\
        Number of layers & 5\\
        Number of heads & 4 \\
        Attention branch dimension & 48 \\
        Value branch dimension & 96 \\
        Hidden Layer MLP & 512 \\
        Attention score dropout & 0.1\\
        Stochastic depth & 0.05\\
        \bottomrule
    \end{tabular}
    \label{tab:arch_attention}
\end{table}

\begin{table}[t]
  \caption{\textbf{Influence of the representations on train time.} We report the iterations per second during training of the model with the QM9 regression task. The timings are averaged over 1 epoch. Batch size is set to 128. We also measure the runtime of the Equiformer~\citep{liao2022equiformer}, which uses computationally more demanding tensor products to achieve equivariance. The models were trained on a single NVIDIA RTX 6000 GPU (CPU: 2x AMD Epyc 7452, 1024 GB RAM).}
  \centering
  \begin{tabular}{@{}lcccc@{}}
    \toprule
    Representation & LoCaFormer $[\text{it}/\text{s}]$ & Diff. to scalar & Equiformer $[\text{it}/\text{s}]$  \\
    \midrule
    Scalar & 4.5 & 0\% & n/a \\
    Cart. Tensor & 3.8 & -16\% & n/a \\
    Irreducible & 3.3 & -27\% & 0.8 \\
    MLP & 4.0 & -11\% & n/a \\
    \bottomrule
  \end{tabular}
  \label{tab:timings_qm9_representations}
\end{table}

\paragraph{Architectural design.} 
In the Equiformer~\citep{liao2022equiformer}, special care is required to use the appropriate specialized linear layers, normalization layers and activation function, whereas in our case a standard MLP and LayerNorm can be employed without breaking equivariance. Beside the LoCaFormer attention block illustrated in Fig.~\ref{fig:attention_mechanisms}, the full LoCaFormer layer consists of the attention block followed by an MLP, together with two vanilla LayerNorm layers, one inserted before the MLP and one before the attention block. Further, we have added skip connections around both modules to improve gradient flow.
The hyperparameters used in our experiments are summarized in Tab.~\ref{tab:arch_attention}. All LoCaFormer models use radial and angular embedding of the relative distance vector with a Bessel function (App.~\ref{app:rad_ang}). The learned local frames are predicted using the procedure described in~\citep{lippmann2024tensor}. As an output head, we use an MLP with hidden dimensions $[512, 128, 32]$, followed by sum-pooling used to aggregate the final node-wise predictions. During training, we use a dropout rate of 0.1 for the attention scores and a stochastic depth of $5\%$ during training.

\paragraph{Tensorial property prediction.} For the tensorial property prediction task on the dataset taken from~\citep{zhang2023learning} the same architecture are used as for property prediction on QM9, except for the following small modifications: The intermediate representations are based on the Cartesian tensor representation ``94x0n + 32x1n + 16x2n'' and the number of message passing layers is reduced to three due to the smaller size of the tensorial property prediction dataset.

\paragraph{Influence of the representations on train time.}\label{ex:influence_representation_time}
In Tab.~\ref{tab:timings_qm9_representations} we report the number of training iterations on QM9 property prediction. While the LoCaFormer with scalar messages still predicts local frames to leverage local canonicalization, it does not perform any (non-trivial) frame-to-frame transitions. Using learned MLP representations (cf.~Sec.~\ref{sec:learning_representations}) or a proper group representation in the frame-to-frame transition introduces a computational overhead but notably improves the predictions in particular on geometric and tensorial targets (see Tabs.~\ref{tab:qm9} and~\ref{tab:tensorial_chemical_properties}). Moreover, the combination of local canonicalization paired with tensorial message passing offers an efficient framework for implementing exact equivariance with optimized standard deep learning building blocks. This can be seen in the direct runtime comparison against the popular Equiformer architecture~\citep{liao2022equiformer}, trained on the same hardware and with the same batch size. Depending on the internal representations used in the message passing of the LoCaFormer, our model is between 4 and 5 times faster than the Equiformer, which employs only irreducible representations; and, unlike our approach, relies on specialized tensor operations to achieve exact equivariance, which are computationally demanding~\citep{passaro2023reducing}.

\subsection{Relation to data augmentation.}\label{app:relation_data_augmentation}

\begin{figure}[t]
\floatconts
  {fig:data_efficiency}
  {\vspace{-0.2cm}\caption{\textbf{Equivariance increases data efficiency compared to data augmentation on QM9.} However, in the low data regime the data augmentation yields superior accuracy.}}
  {%
    \subfigure[Data efficiency plot with $\epsilon_{\text{LUMO}}$ as target]{\label{fig:image-a}%
      \includegraphics[width=0.42\linewidth]{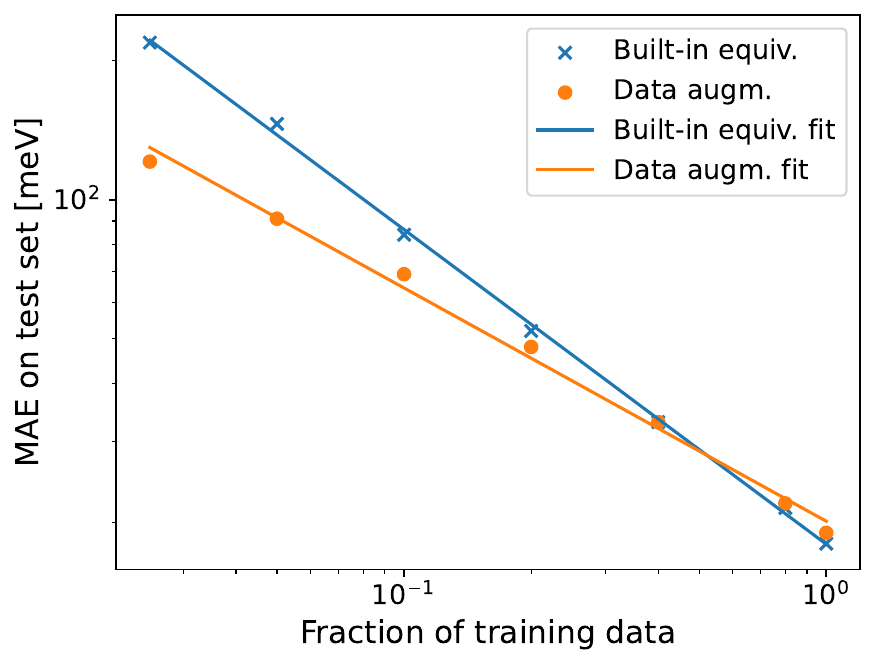}}%
    \qquad
    \subfigure[Data efficiency plot with $\mu$ as target]{\label{fig:image-b}%
      \includegraphics[width=0.42\linewidth]{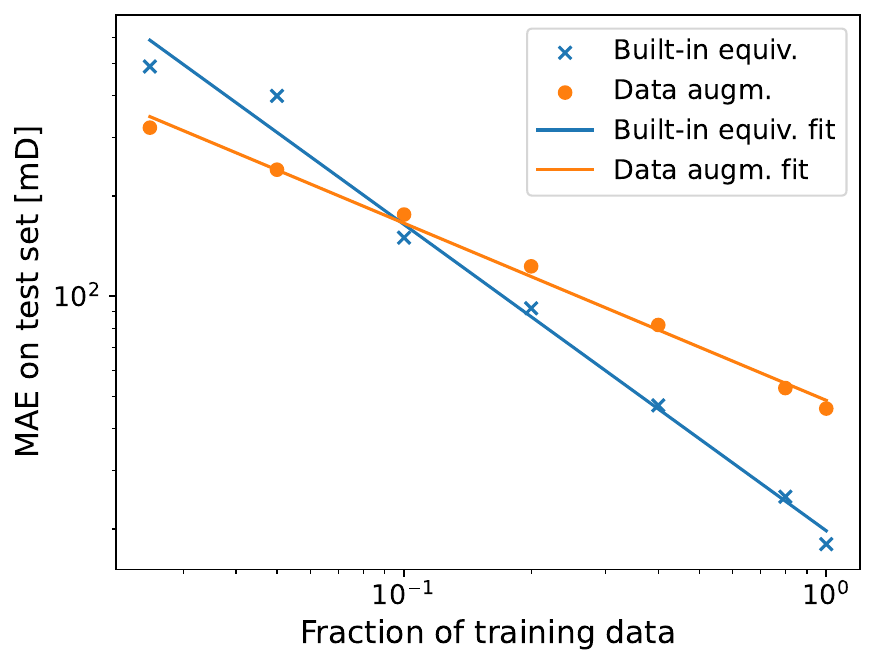}}
  }
\end{figure}

\noindent The framework of local canonicalization makes it straightforward to compare data augmentation (by choosing the same random local frame for each node) with built-in equivariance. Equivariant models are often considered more data-efficient because they do not need to spend model capacity or additional data to learn symmetries~\citep{batzner20223}, implying faster performance gains as training data increases~\citep{hestness2017deep}. To test this, we trained the best equivariant model (according to Tab.~\ref{tab:qm9}) and its data-augmented counterpart (same architecture and hyperparameters) on varying fractions of the training set and compared test accuracies. We evaluated two targets: the norm of the magnetic dipole moment $\mu$, where tensorial messages improve performance, and $\epsilon_{\text{LUMO}}$, where tensorial messages have little effect. In both cases, the model with built-in equivariance exhibits steeper error–data scaling, cf.~Fig.~\ref{fig:data_efficiency}, confirming higher data efficiency~\citep{hestness2017deep}. However, interestingly, the equivariant model did not consistently achieve lower error across all data fractions: in the low-data regime, the augmented model sometimes outperformed it. This challenges the common assumption that exact equivariance would dominate in low-data settings.

\end{document}